\documentclass[journal]{IEEEtran}
\usepackage{times}  
\usepackage{helvet} 
\usepackage{courier}  
\usepackage[hyphens]{url}  
\usepackage{graphicx} 
\usepackage{graphicx}  
\usepackage{subcaption}
\usepackage{booktabs}
\usepackage{amsmath, bm}
\usepackage{amsfonts}
\usepackage{verbatim}
\usepackage{multirow}
\usepackage{multicol}

\usepackage{soul}
\usepackage{color, xcolor}

\definecolor{mycolor}{RGB}{0,0,0}
\DeclareCaptionFont{mycolor}{\color{mycolor}}

\usepackage{algorithm}
\usepackage{algorithmic}
\usepackage{hyperref}
\hypersetup{
colorlinks=true,
linkcolor=blue,
anchorcolor=blue,
citecolor=blue
}

\usepackage{array}
\usepackage{colortbl}

\makeatletter
\newcommand{\removelatexerror}{\let\@latex@error\@gobble}
\makeatother

\begin{document}

\title{Memorizing Complementation Network for Few-Shot Class-Incremental Learning}

\author{Zhong~Ji,~\IEEEmembership{Senior~Member,~IEEE,}
		Zhishen~Hou,
        Xiyao~Liu,~\IEEEmembership{Graduate~Student~Member,~IEEE,}
        Yanwei~Pang,~\IEEEmembership{Senior~Member,~IEEE,}
        Xuelong~Li,~\IEEEmembership{Fellow,~IEEE}

\thanks{Manuscript received xxx xx, 2022; revised xxx xx, 2022.}
\thanks{This work was supported by the National Natural Science Foundation of China (NSFC) under Grants 62176178.}
\thanks{Z. Ji, Z. Hou, X. Liu*(corresponding author), and Y. Pang are with the School of Electrical and Information Engineering; Tianjin Key Laboratory of Brain-inspired Intelligence Technology, Tianjin University, Tianjin 300072, China (e-mails: jizhong@tju.edu.cn; zshou@tju.edu.cn; xiyaoliu@tju.edu.cn; pyw@tju.edu.cn).} 
\thanks{X. Li is with School of Artificial Intelligence, Optics and Electronics (iOPEN), Northwestern Polytechnical University, Xi'an 710072, P.R. China, (e-mail: li@nwpu.edu.cn)}}

\markboth{Journal of IEEE TRANSACTIONS ON IMAGE PROCESSING,~Vol.~nn, No.~n, August~20nn}%
{Shell \MakeLowercase{\textit{$et\, al.$}}: Bare Demo of IEEEtran.cls for IEEE Journals}

\maketitle

\begin{abstract}
Few-shot Class-Incremental Learning (FSCIL) aims at learning new concepts continually with only a few samples, which is prone to suffer the catastrophic forgetting and overfitting problems. The inaccessibility of old classes and the scarcity of the novel samples make it formidable to realize the trade-off between retaining old knowledge and learning novel concepts. Inspired by that different models memorize different knowledge when learning novel concepts, we propose a Memorizing Complementation Network (MCNet) to ensemble multiple models that complements the different memorized knowledge with each other in novel tasks. Additionally, to update the model with few novel samples, we develop a Prototype Smoothing Hard-mining Triplet (PSHT) loss to push the novel samples away from not only each other in current task but also the old distribution. Extensive experiments on three benchmark datasets, e.g., CIFAR100, miniImageNet and CUB200, have demonstrated the superiority of our proposed method.
\end{abstract}

\begin{IEEEkeywords}
Few-Shot Learning, Class-Incremental Learning, Ensemble Learning, Memorizing Complementation.
\end{IEEEkeywords}

%
\IEEEpeerreviewmaketitle

\section{Introduction}
\IEEEPARstart{D}{eep} Convolutional Neural Networks have achieved significant success in a large amount of vision tasks depending on multitudinous data and unprecedented computational resources. However, the deep learning-based methods fail to recognize untrained classes. The model has to be retrained with both old classes and novel classes, which is time-consuming and expensive. In this case, Class-Incremental Learning (CIL) is extensively studied to update the model on-the-fly only with novel samples in each incremental task while maintaining the knowledge of old classes. However, it is difficult to acquire and annotate the novel samples in some practical applications, which leads to the inapplicability of CIL methods demanding for abundant novel samples. Therefore, some work explores to study the CIL in the low-data scenario, called Few-Shot Class-Incremental Learning (FSCIL) \cite{TOPIC,SPPR,CEC,zhao2021mgsvf}.

Different from traditional CIL, FSCIL deploys incremental tasks continually with only a few novel samples, which results in not only the catastrophic forgetting for old classes but also the overfitting for new concepts. Thus, it is essential to address these two problems simultaneously in FSCIL. To this end, recent studies \cite{SPPR,CEC,F2M} explored to learn the transferable features with abundant base classes and limit the changes of model parameters in incremental tasks. For example, Zhu $et\, al.$ \cite{SPPR} randomly selected episodes as the simulated incremental tasks to train the model with base samples to make sure the features extracted from the model are adaptive to various incremental tasks. Similarly, Zhang $et\, al.$ \cite{CEC} trained a graph-based classifier with a pseudo incremental-learning paradigm.  Shi $et\, al.$ \cite{F2M} explored to search the flat local minima with base samples so that the model can be updated within the flat region to reduce the knowledge forgetting. These approaches all employed a single model to overcome the problems mentioned above. However, it is limited of the ability to resist catastrophic forgetting with only a single model.

\begin{figure}[t]
    \centering
    \includegraphics[width=\linewidth]{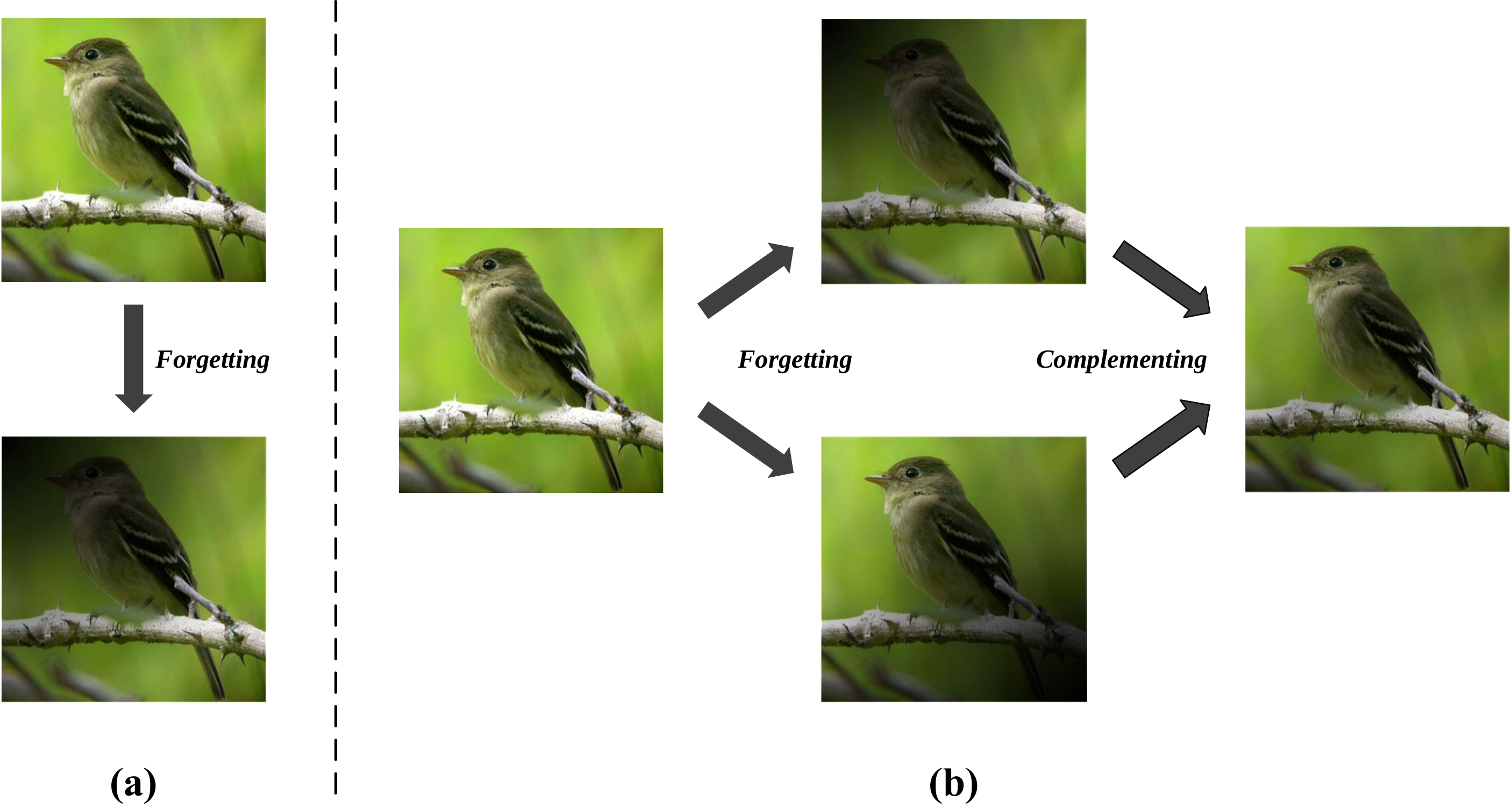}    
    \caption{An illustration of the memorizing complementation. (a) represents the memorized knowledge for the bird after forgetting old knowledge and learning new concepts with a single model. (b) denotes that ensembling multiple models memorizing different knowledge can realize the complementation and still remain the discriminant features of the bird.}
    \label{fig:visual}
\end{figure}

Actually, a single model largely focuses on one-side representations while different models have distinct emphasis \cite{shao2021mhfc,chowdhury2021few}. Thus, different models may memorize different knowledge when learning novel concepts continually. As show in Fig. \ref{fig:visual}, for recognizing a bird, the beak, wings, claws and tails are crucial parts for a model to concentrate on. When updated with novel classes, some models may memorize the knowledge of the beak or wings, while other models may memorize the claws or tails. This means that the knowledge retained by these models are complemented with each other and the ensembled models still have the ability of recognizing the birds. To this end, we explore to ensemble multiple models to realize the memorizing complementation.

In order to force different models to memorize different knowledge when learning novel samples, it is crucial to obtain the models with both discrimination and diversity. There are mainly two types of approaches to improve the diversity of different models: 1) constructing models with different structures, and 2) training models with different tasks. In this paper, we apply the above two approaches. Firstly, recent studies \cite{d2021convit,Yuan_2021_ICCV,ge2021revitalizing} have explained that the CNN is good at extracting the local features but difficult to capture the global representations, while the Visual Transformer \cite{dosovitskiy2020vit} can obtain the long-distance dependencies but is weak in capturing detail information. Therefore, we utilize a CNN-structure and a Transformer-structure to obtain local-focus features and global-focus representations respectively. Additionally, considering that the model focuses on the different parts, e.g., the contour information or the texture information of samples when dealing with different tasks, we explore to train multiple models with various tasks, e.g., the traditional classification task and the cross-modal classification task. 

Meanwhile, considering the novel samples are inevitably overlapped with the old classes in the embedding space, which is dubbed as the ambiguities problem \cite{LUCIR}, it is essential to update the model in each task. Recent approaches \cite{yu2020sdc,FFSL} generally employ a metric-based loss with a knowledge distillation regularization to fine-tune the model. However, the original triplet loss cannot construct the relations between old and novel classes. Besides, since the old samples are unavailable in novel sessions, it is difficult to realize the trade-off between learning new knowledge and retaining old knowledge, especially in the few-shot scenario. Therefore, we constrain the similarity among different samples of the novel classes and the old classes simultaneously and only fine-tune a small part of the whole model.

Specifically, we propose a Memorizing Complementation Network (MCNet) to ensemble multiple complemental models to alleviate the catastrophic forgetting problem of the FSCIL. The MCNet consists of  a Structure-Wise Complementation (SWC) strategy and a Task-Wise Complementation (TWC) strategy respectively to acquire various embedding networks. The SWC contains a CNN head and a Transformer head to extract different features, while the TWC trains models through a traditional classification task with a cosine classifier and a cross-modal classification task with a metric-based classifier. We also utilize an Attention Regularization (AR) to further improve the diversity of different models. To update the models in novel tasks, we propose a Prototype Smoothing Hard-mining Triplet (PSHT) loss that not only constrains the similarities among different samples in current session but also pushes the novel samples and the old distribution away.

Our highlights are summarized in three folds:
\begin{enumerate}
    \item We propose a Memorizing Complementation Network (MCNet), which consists of the Structure-Wise Complementation (SWC) and the Task-Wise Complementation (TWC) strategy, to train multiple embedding networks that can complement the memorized knowledge with each other. Additionally, we employ an Attention Regularization (AR) to further improve the ability of the complementaion. To the best of our knowledge, this is the first work to introduce ensemble learning idea into FSCIL.
    \item To alleviate the catastrophic forgetting and overfitting problems when updating the model, we employ a Prototype Smoothing Hard-mining Triplet (PSHT) loss that simultaneously constrains the similarity of old classes and novel classes.
    \item We conduct experiments on three benchmark datasets, i.e., CIFAR100, miniImageNet and CUB200. The experimental results demonstrate the effectiveness of our method that outperforms the state-of-the-art approaches by a large margin.
\end{enumerate}

\begin{figure*}[t]
    \centering
    \includegraphics[width=0.95\linewidth]{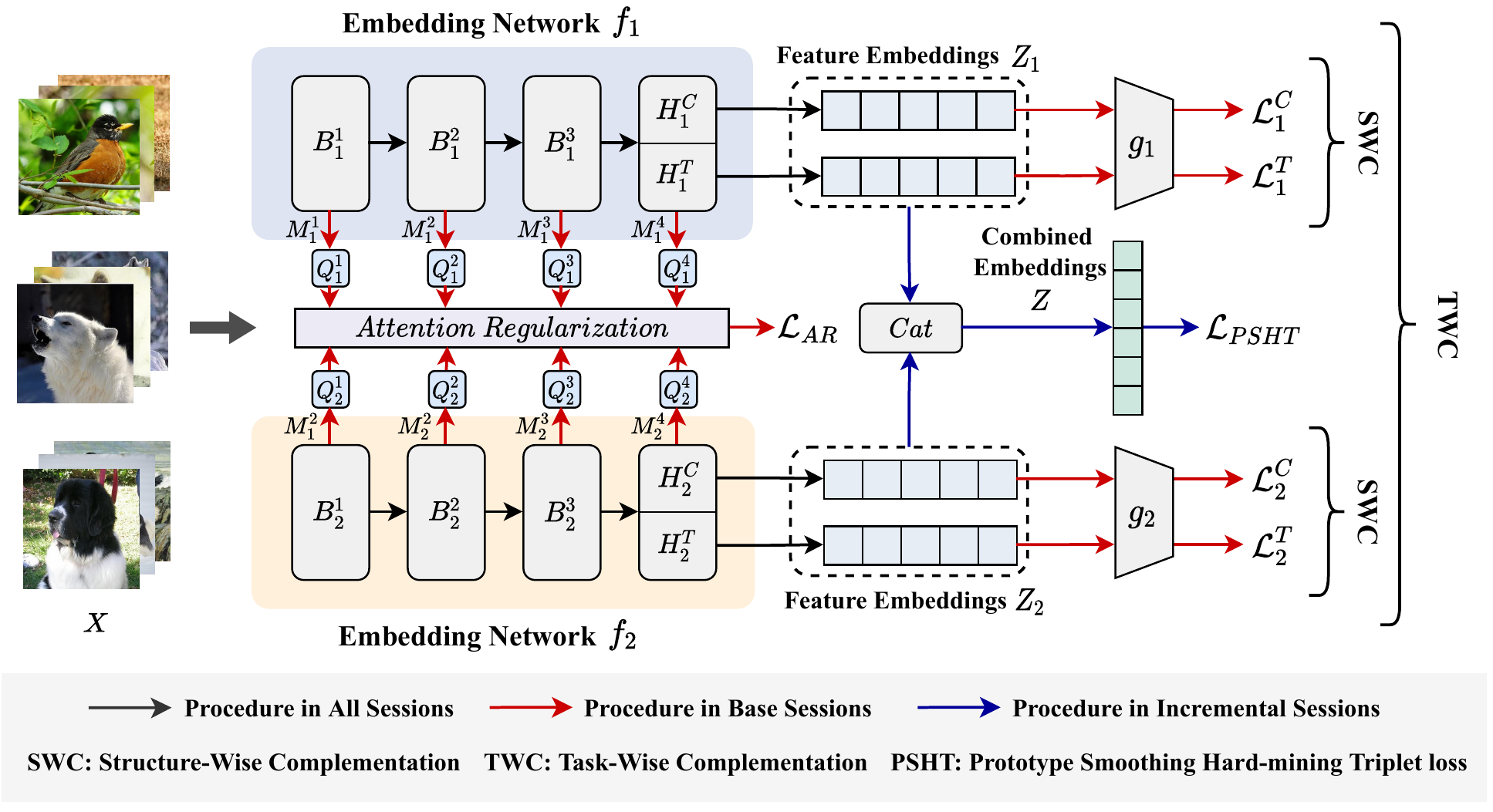}
    \caption{The framework of our proposed MCNet. In the base session, we train different models that complements the memorized knowledge with each other. In the incremental sessions, we employ the Prototype Smoothing Hard-mining Triplet (PSHT) loss to update the model. The red line represents the training procedure in base session and the blue line denotes that in incremental sessions.}
    \label{fig:framework}
\end{figure*}

\section{Related Work}

\subsection{Few-Shot Learning}
Few-Shot Learning aims at learning a model that can fast adapt to the novel concepts with only a few labeled samples, which is widely studied in recent years. Current FSL methods are mainly divided into two types: 1) The meta-learning-based FSL, which includes the optimization-based methods \cite{finn2017model-agnostic,rusu_meta-learning_2019}, the metric-based methods \cite{vinyals2016matching,Snell2017Prototypical,sung2018learning,mapnet}, and the data-augmentation-based methods \cite{wang2018low-shot,li2020adversarial}. The first group of methods tends to learn the sub-optimal parameters for each task as the initial parameters that can be quickly adapted to the novel classes. The metric-based methods learn a generalizable embedding space and measure the similarity among feature embeddings to classify the novel samples. As for the data-augmentation-based methods, the main idea is to generate high-quality features to alleviate the lack of labeled data. 2) The transfer learning-based FSL, which trains the deep embedding networks with base samples in a standard classification paradigm and transfer the prior knowledge into novel classes. Recent work \cite{chen2019a,tian2020rethinking,rizve2021exploring} has outperformed the sophisticated meta-learning methods. For example, Chen $et\, al.$ \cite{chen2019a} conducted experiments to demonstrate that a distance-based classifier that is competitive to the meta-learning methods. After that, Tian $et\, al.$ \cite{tian2020rethinking} introduced the self-distillation into FSL to enhance the feature embeddings. Rizve $et\, al.$ \cite{rizve2021exploring} followed this paradigm and proposed a novel training mechanism that simultaneously enforces equivariance and invariance to a general set of geometric transformations.

\subsection{Incremental Learning} 
Incremental learning focuses on the problem of learning new concepts continually with a sequence of data without forgetting the old knowledge. Generally, the incremental learning includes three types: multi-task incremental learning \cite{chaudhry2018riemannian,nguyen2018variational,riemer2019learning}, multi-domain incremental learning \cite{shin2017continual,zenke2017continual} and multi-class incremental learning \cite{icarl,EEIL,LUCIR,yu2020sdc}. Rebuffi $et\, al.$ \cite{icarl} preserved valuable samples of old classes as exemplars that are distilled to mitigate the forgetting and utilized a nearest-neighbor classifier to learn the new samples. Castro $et\, al.$ \cite{EEIL} introduced an end-to-end framework that combines a classification loss and a knowledge distillation loss. Focusing on the imbalance problem between the old and new classes, Hou $et\, al.$ \cite{LUCIR} explored to learn a unified classifier that integrates cosine normalization, less-forget constraint and inter-class separation. Yu $et\, al.$ \cite{yu2020sdc} proposed a semantic drift compensation to estimate the new class center after updating the model in the incremental sessions. In this paper, we pay only attention on the multi-class incremental learning that is the most challenge and practical.

\subsection{Few-Shot Class-Incremental Learning}
FSCIL \cite{TOPIC,SPPR,CEC,F2M,cheraghian2021semantic,cheraghian2021synthesized,wang2020meta} is proposed recently to simultaneously address the catastrophic forgetting problem in incremental learning and the overfitting problem in few-shot learning, which is more practical in real world. As the pioneering work, Tao $et\, al.$ \cite{TOPIC} utilized a Neural Gas (NG) network to preserve the topology of the feature embeddings of the old classes. Afterwards, Zhu $et\, al.$ \cite{SPPR} imitated the meta-learning paradigm and employed a random episode selection strategy to force the features adaptive to various simulated incremental sessions. In each simulated sessions, they introduced a Self-Promoted Prototype Refinement mechanism (SPPR) to enhance the prototypes by constructing the relationships between old classes and new samples. Cheraghian $et\, al.$ \cite{cheraghian2021semantic} utilized the semantic information of each category for classification and to facilitate the knowledge distillation that is essential for alleviating the catastrophic forgetting problem. In \cite{cheraghian2021synthesized}, the authors employed a mixture of subspaces to align the visual and semantic embeddings and synthesized novel features to address the overfitting problems. Recently, some studies \cite{CEC,F2M} discovered that an intransigent model that only trains on base classes outperforms state-of-the-art methods. Based on this, Zhang $et\, al.$ \cite{CEC} proposed a Continually Evolved Classifier (CEC) that employs a graph model to obtain the session-wise classifiers that are trained with a pseudo incremental learning paradigm. Shi $et\, al.$ \cite{F2M} explored to search the flat local minima of the base training objective function so that the model parameters are updated within the flat region in the incremental sessions, which efficiently overcomes the catastrophic forgetting problem. In this paper, we also focus on the base session with abundant samples to address the FSCIL, while the main idea is to train a model that can complement the memorized knowledge in incremental sessions.

\subsection{Ensemble Learning}
Ensemble learning is well known for its effectiveness to reduce the variance of classifiers and improve the performance. The ensemble methods are widely employed in variety of machine learning tasks \cite{shao2021mhfc,chowdhury2021few,zhang2012ensemble,ding2017low,dvornik2019diversity,zhou2021domain}. In recent years, considering the large variance in the low data scenario, some studies have proved that the few-shot learning benifits significantly from the ensemble method. For example, Dvornik $et\, al.$ \cite{dvornik2019diversity} first introduced the ensemble method into the FSL, which encourages the cooperation and diversity of the ensemble networks. Shao $et\, al.$ \cite{shao2021mhfc} also ensembled multiple models to enhance the feature embeddings. In addition, they proposed a multi-head feature collaboration (MHFC) algorithm to capture more discriminative information. Chowdhury $et\, al.$ \cite{chowdhury2021few} conducted abundant experiments to show that an ensemble of pre-trained embedding networks with a classifier can well address the FSL. They discovered that the diversity of the model may be likely more important than the large number of training data. Motivated by this discovery, we consider the model diversity from the perspective of incremental learning and explore to ensemble diverse models to mitigate the catastrophic forgetting problem.


\section{Memorizing Complementation Network}

\subsection{Preliminary}

The Few-Shot Class-Incremental Learning (FSCIL) aims at learning to recognize new categories with only few samples continually while maintaining the ability of recognizing old classes. Specifically, there is a sequence of training sets $\{\mathcal{D}_1, \mathcal{D}_2, \cdots, \mathcal{D}_T\}$ in FSCIL, where $\mathcal{D}_t  = \{\textbf{x}_i,\textbf{y}_i, \textbf{a}_i\}^n_{i=0},t \in \{1, 2, \cdots, T\}$ is the training set in the $t$-th session, where $\textbf{x}_i,\textbf{y}_i, \textbf{a}_i$ represent the $i$-th image, the corresponding label and the semantic vector, respectively. Notably, the classes are disjoint among all sessions, i.e., $\mathcal{C}_i \cap  \mathcal{C}_j = \emptyset, \forall i,j$ and $i \neq j$. In the $t$-th session, the model is trained with only $\mathcal{D}_t$ and evaluated with all classes $\{\mathcal{C}_1, \mathcal{C}_2, \cdots, \mathcal{C}_t\}$ that have been encountered yet. Different from the traditional class-incremental learning, there are adequate training data in the first session (called base session), while the following sessions (called incremental sessions) only have a few samples in each class. 

In this paper, we propose a Memorizing Complementation Network (MCNet) to ensemble multiple embedding networks that complement the remained knowledge with each other. The main framework of our proposed method is presented in Fig. \ref{fig:framework}. Rather than utilizing a trainable classifier, we employ a prototype-based classifier without any parameters to mitigate the overfitting problem following \cite{CEC,F2M,FFSL}. Specifically, we calculate the centroid of each class as prototypes and utilize the similarity between each samples and the prototypes for classification, which is defined as:
\begin{equation}
  \hat{\textbf{y}}_i = \mathop{\arg\max}\limits_{c \in \cup_1^t \mathcal{C}^t} sim( f(\textbf{x}_i),\textbf{p}_c ) ,
\end{equation}
where $f$ denotes the embedding network, $\textbf{p}_c$ is the prototypes of class $c$ and $sim(\cdot, \cdot)$ is the similarity function, e.g., cosine similarity. 


\subsection{Memorizing Complementation Network}

To make sure the different embedding networks can complement each other effectively, it is essential to improve the diversity of these networks while maintain the discrimination. To this end, we propose a  Structure-Wise Complementation (SWC) strategy and a Task-Wise Complementation (TWC) strategy respectively.

\subsubsection{Structure-Wise Complementation}

One intuitive idea to improve the model diversity is directly utilizing different structures to construct the embedding networks. The convolution operation is considered to be good at extracting the local features but short in capturing the global representations, while the self-attention in Visual Transformer can obtain the long-distance dependencies but weak in capturing detail information. Thus, we employ a CNN structure as one head and a Transformer structure as the other head to obtain local-focus features and global-focus features respectively. Rather than aligning these two features, we regard them as two view features of one instance and classify these two features respectively to make sure they are different but both discriminative, which improves the model diversity and realizes the complementation between local and global features. 

Specifically, the CNN head is constructed with several residual blocks, while the Transformer head consists of some consecutive attention blocks. The architecture of the attention block follows \cite{srinivas2021bottleneck}, where each block contains two Multi-Layer Perception (MLP) layers with one Multi-Head Self-Attention (MHSA) layer between them. The MHSA is defined as:
\begin{equation}
  Attention(\bm{q},\bm{k},\bm{v})=softmax(\frac{\bm{q}\bm{e}^T+\bm{q}\bm{k}^T}{\sqrt{d_k}})\bm{v}, 
\end{equation}
where $\bm{q}, \bm{k}, \bm{v}$ are the query, key, value features mapped from the input feature map with three different MLP layers, and $\bm{e}$ is the position encoding and $d_k$ is the dimensionality of the query and key features.	

The outputs of these two heads are 2D feature maps with the same size, which are then flatten into feature vectors with an average pooling operation. We denote these two features as $\textbf{z}^C=h^C (B(\textbf{x}))$ and $\textbf{z}^T=h^T (B(\textbf{x}))$, where $B(\cdot)$ is the base embedding network and $h^C$ and $h^T$ are CNN head and Transformer head respectively. Both features are then input into a classifier when training in the base session as follows:
\begin{equation} \label{Loss_C}
  \mathcal{L}^C = \mathcal{L}_{CE} (g(\textbf{z}^C; \phi), \textbf{y}) ,
\end{equation}
\begin{equation} \label{Loss_T}
  \mathcal{L}^T = \mathcal{L}_{CE} (g(\textbf{z}^T; \phi), \textbf{y}) ,
\end{equation}
where $g(\cdot; \phi)$ represents a classifier. After the base training, we ensemble the multi head features for evaluation and fine-tune in incremental sessions.

\subsubsection{Task-Wise Complementation}

Since one model can only extract one-side representations while different models have distinct emphases \cite{shao2021mhfc,chowdhury2021few}, it is necessary to utilize multiple models to obtain various and complementary knowledge. Thus, we explore to train multiple models with different tasks. Specifically, one model is trained with a vanilla cosine classifier constructed by a fully-connected layer without bias, while the other model realizes the cross-modal classification by calculating the similarity between the features and the semantic vectors of all categories. The objective functions of these two models are:
\begin{align}
  \mathcal{L}_1 & = \mathcal{L}_{CE} (\cos(\textbf{z}_1, \phi), \textbf{y}) , \\
  \mathcal{L}_2 & = \mathcal{L}_{CE} (\cos(\textbf{z}_2, s (\textbf{a}; \psi)), \textbf{y}) ,
\end{align}
where $\textbf{z}_1$ represents $\textbf{z}_1^C=h^C_1 (B_1 (\textbf{x}))$ or $\textbf{z}_1^T=h^T_1 (B_1 (\textbf{x}))$ and $\textbf{z}_2$ denotes $\textbf{z}_2^C=h^C_2 (B_2 (\textbf{x}))$ or $\textbf{z}_2^T=h^T_2 (B_2 (\textbf{x}))$, 
$\phi$ are the parameters of the cosine classifier, $s$ is the semantic embedding network and $\psi$ are the corresponding parameters. In this case, one model focuses on the visual difference such as structure or texture while the other model pays more attention on the semantic difference, which makes sure the models memorize different knowledge in incremental sessions. 

\subsubsection{Attention Regularization}
To further improve the diversity between different models, it is intuitive to employ a regularization to reduce the similarity between the output features. However, it may hurt the discrimination of the model when directly regularizing the output features, e.g., reducing the cosine similarity between these two features. Therefore, we develop an Attention Regularization (AR) to urge different models to focus on different parts, which prompts the output features to be more different and complementary. The attention regularization term is defined as:
\begin{equation}
  \mathcal{L}_{AR} = \frac{1}{L} \sum_j^L \frac{\textbf{Q}^j_1 \cdot \textbf{Q}^j_2}{\Vert \textbf{Q}^j_1 \Vert_2 \cdot \Vert \textbf{Q}^j_2 \Vert_2 } , 
\end{equation}
where $L$ denotes the number of layer blocks and $j$ is the current layer, and $\textbf{Q}$ is the activation-based spatial attention maps as in \cite{zagoruyko2016paying}, which is defined as:
\begin{equation}
  \textbf{Q} =  \sum_c \Vert \textbf{M}_c \Vert_2^2 , 
\end{equation}
where $\textbf{M}_c$ is the feature map of channel $c$ and $\textbf{Q}$ is flattened to the vectorized form.

We employ the SWC, the TWC and the AR strategies simultaneously to make sure the embedding networks complement the memorized knowledge with each other. Each model contains two heads and outputs two types of feature embeddings. Therefore, we can obtain four types of output features totally, which are denoted as $\textbf{z}^C_1=h^C_1 (B_1 (\textbf{x}))$, $\textbf{z}^T_1=h^T_1 (B_1 (\textbf{x}))$, $\textbf{z}^C_2=h^C_2 (B_2 (\textbf{x}))$ and $\textbf{z}^T_2=h^T_2 (B_2 (\textbf{x}))$. The total loss in the base session is defined as:
\begin{equation} \label{Loss_base}
  \mathcal{L}_{base} = \mathcal{L}^C_1 + \mathcal{L}^T_1 + \mathcal{L}^C_2 + \mathcal{L}^T_2 + \alpha \mathcal{L}_{AR}. 
\end{equation}
During the evaluation stage, we first compose these features of training samples to calculate the composed prototypes of all classes, and then utilize the cosine similarity between the composed features of test data and the composed prototypes to finish the classification. In this paper, we employ the concatenation operation for feature composition and denote the composed features as $\textbf{z}=F(\textbf{x})$ for convenience.

\begin{figure}[t]
    \centering
    \includegraphics[width=\linewidth]{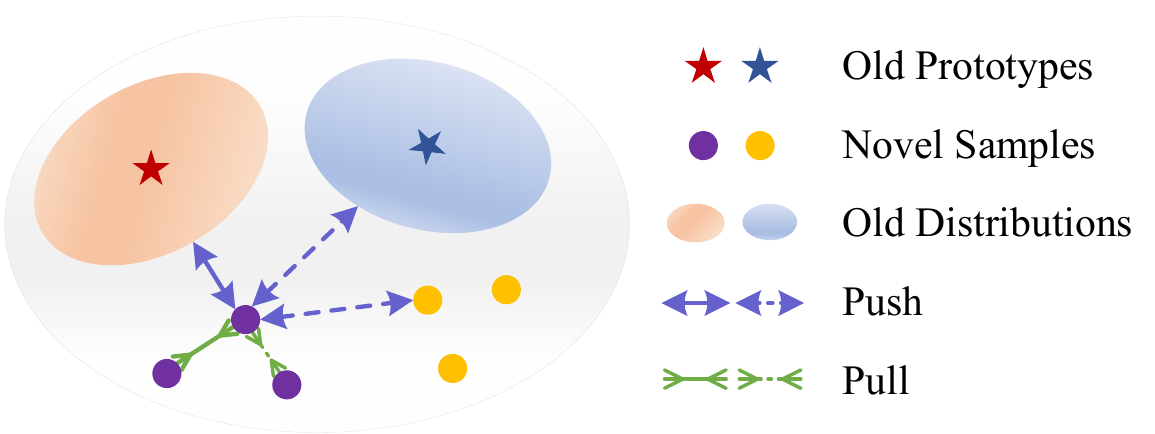}    
    \caption{An illustration of the Prototype Smoothing Hard-mining Triplet Loss. The solid arrows represent the hard-mining strategy, while the dash arrows denote the implicit operations between easy samples.}
    \label{fig:PSTH}
\end{figure}

\subsection{Prototype Smoothing Hard-mining Triplet Loss}

Since the novel samples are usually overlapped with the old samples in feature space, it is essential to fine-tune the model in each session. Unfortunately, there are only several samples in novel sessions and all previous training sets are inaccessible so that the overfitting and the catastrophic forgetting are still inevitable. To alleviate these problems, we present a Prototype Smoothing Hard-mining Triplet (PSHT) loss function to update the MCNet. 

Specifically, we assume the old samples follow the Gaussian Distribution with the corresponding prototypes as their average values. Based on this assumption, the novel samples should be far away from the distribution space of the old classes to ensure that all samples are classified correctly. To this end, we first calculate and store the diagonal covariance $diag(\sigma^2)$ of each old class, which is utilized to smooth the prototypes and restore the distribution. To implement the operation mentioned above, we regard the vectors sampled from the distribution as the feature embeddings of the old samples, which is defined as:
\begin{equation} \label{guassian}
  \tilde{\textbf{z}}_k^c = \textbf{p}_c + \eta * diag(\sigma), 
\end{equation}
where $\tilde{\textbf{z}}_k^c$ is the $k$-th sample of class $c$, $\textbf{p}_c$ is the prototypes of class $c$ and $\eta \sim \mathcal{N}(0, 1)$ is a standard gaussian noise. Then, we combine these old samples and current samples as training set to update the model with a hard-mining triplet loss, which is define as: 
\begin{equation} \label{Loss_trip}
  \mathcal{L}_{PSHT} = \mathop{\max} (0, \Vert \textbf{z}-\textbf{z}^+_{max} \Vert_2 - \Vert \textbf{z}-\textbf{z}^-_{min} \Vert_2), 
\end{equation}
where $\textbf{z}_{max}^+$ is the farthest embedding that belongs to the same class and $\textbf{z}_{min}^-$ is the nearest embedding that belongs to the different classes. Since the old feature embeddings are directly sampled from the gaussian distribution, we do not utilize them as the anchor points. Thus, $\textbf{z}$ in Eq. (\ref{Loss_trip}) only contains the feature embeddings of current samples. The illustration for the PSHT loss is shown in Fig. \ref{fig:PSTH}. In addition, considering the shallow layers of neural network tend to learn the lower features, e.g., edge, color or texture, which are usually general for all images and more transferable, we choose to fixed the shallow layers and only fine-tune the deep layers in the incremental sessions. 

We also apply a knowledge distillation regularization term to maintain the old knowledge as follows:
\begin{equation} \label{Lr}
  \mathcal{L}_{KD} = \Vert \textbf{z}^t - \textbf{z}^{t-1} \Vert_2. 
\end{equation}
Therefore, the total loss in the incremental sessions is:
\begin{equation} \label{Loss_all}
  \mathcal{L}_{novel} = \mathcal{L}_{PSHT} + \lambda \mathcal{L}_{KD}. 
\end{equation}

\section{Experiments}

\setlength{\tabcolsep}{4pt}
\begin{table*}[t]
\begin{center}
    \caption{Comparison with the state-of-the-art methods on CIFAR100 dataset.}
    \label{table:cifar}
    \setlength{\tabcolsep}{2mm}
	\renewcommand\arraystretch{1.1}
    \begin{tabular}{lccccccccccc}
        \hline\noalign{\smallskip}
        \multirow{2}*{\textbf{Methods}}     & \multicolumn{9}{c}{\textbf{Acc. in each session (\%)}}    \\ \cline{2-10}
        \noalign{\smallskip}
		~                          & 0 &  1  & 2 & 3 & 4 & 5 & 6 & 7 & 8      \\
        \hline
        \noalign{\smallskip}
        iCaRL \cite{icarl}     & 64.10 & 53.28 & 41.69 & 34.13 & 27.93 & 25.06 & 20.41 & 15.48 & 13.73   \\
        EEIL \cite{EEIL}   & 64.10 & 53.11 & 43.71 & 35.15 & 28.96 & 24.98 & 21.01 & 17.26 & 15.85   \\
        LUCIR \cite{LUCIR}       & 64.10 & 53.05 & 43.96 & 36.97 & 31.61 & 26.73 & 21.23 & 16.78 & 13.54    \\
        TOPIC \cite{TOPIC}     & 64.10 & 55.88 & 47.70 & 45.16 & 40.11 & 36.38 & 33.96 & 31.55 & 29.37  \\
        SKD \cite{cheraghian2021semantic}   & 64.10 & 57.00 & 50.01 & 46.00 & 44.00 & 42.00 & 39.00 & 37.00 & 35.00 \\
        SPPR \cite{SPPR}   & 63.97 & 65.86 & 61.31 & 57.60 & 53.39 & 50.93 & 48.27 & 45.36 & 43.32 \\
        SFMS \cite{cheraghian2021synthesized}   & 62.57 & 57.00 & 56.70 & 52.00 & 50.60 & 48.80 & 45.00 & 44.00 & 41.64   \\
        \noalign{\smallskip}
        \hline
        \noalign{\smallskip}
        FFSL \cite{FFSL}   & 72.28 & 63.84 & 59.64 & 55.49 & 53.21 & 51.77 & 50.93 & 48.94 & 46.96  \\
        CEC \cite{CEC}   & 73.07 & 68.88 & 65.26 & 61.19 & 58.09 & 55.57 & 53.22 & 51.34 & 49.14  \\
        F2M \cite{F2M}   & 71.45 & 68.10 & 64.43 & 60.80 & 57.76 & 55.26 & 53.53 & 51.57 & 49.35  \\
        \noalign{\smallskip}
        \hline
        \noalign{\smallskip}
        \textbf{MCNet (Ours)}       & \textbf{73.30} & \textbf{69.34} & \textbf{65.72} & \textbf{61.70} & \textbf{58.75} & \textbf{56.44} & \textbf{54.59} & \textbf{53.01} & \textbf{50.72}   \\
        \noalign{\smallskip}
        \hline
        
    \end{tabular}
\end{center}
\end{table*}
\begin{table*}[!t]
\begin{center}
    \caption{Comparison with the state-of-the-art methods on miniImageNet dataset.}
    \label{table:mini}
    \setlength{\tabcolsep}{2mm}
	\renewcommand\arraystretch{1.1}
    \begin{tabular}{lccccccccccc}
        \hline\noalign{\smallskip}
        \multirow{2}*{\textbf{Methods}}     & \multicolumn{9}{c}{\textbf{Acc. in each session (\%)}}    \\ \cline{2-10}
        \noalign{\smallskip}
		~                          & 0 &  1  & 2 & 3 & 4 & 5 & 6 & 7 & 8      \\
        \hline
        \noalign{\smallskip}
        iCaRL \cite{icarl}     & 61.31 & 46.32 & 42.94 & 37.63 & 30.49 & 24.00 & 20.89 & 18.80 & 17.21   \\
        EEIL \cite{EEIL}      & 61.31 & 46.58 & 44.00 & 37.29 & 33.14 & 27.12 & 24.10 & 21.57 & 19.58   \\
        LUCIR \cite{LUCIR}    & 61.31 & 47.80 & 39.31 & 31.91 & 25.68 & 21.35 & 18.67 & 17.24 & 14.17    \\
        TOPIC \cite{TOPIC}     & 61.31 & 50.09 & 45.17 & 41.16 & 37.48 & 35.52 & 32.19 & 29.46 & 24.42  \\
        SKD \cite{cheraghian2021semantic}   & 61.31 & 58.00 & 53.00 & 50.00 & 48.00 & 45.00 & 42.00 & 40.00 & 39.00 \\
        SPPR \cite{SPPR}   & 61.45 & 63.80 & 59.53 & 55.53 & 52.50 & 49.60 & 46.69 & 43.79 & 41.92 \\
        SFMS \cite{cheraghian2021synthesized}    & 61.40 & 59.80 & 54.20 & 51.69 & 49.45 & 48.00 & 45.20 & 43.80 & 42.10  \\
        \noalign{\smallskip}
        \hline
        \noalign{\smallskip}
        FFSL \cite{FFSL}   & 72.08 & 59.04 & 53.75 & 51.17 & 49.11 & 47.21 & 45.35 & 44.06 & 43.65  \\
        CEC \cite{CEC}   & 72.00 & 66.83 & 62.97 & 59.43 & 56.70 & 53.73 & 51.19 & 49.24 & 47.63  \\
        F2M \cite{F2M}   & 72.05 & 67.47 & 63.16 & 59.70 & 56.71 & 53.77 & 51.11 & 49.21 & 47.84  \\
        \noalign{\smallskip}
        \hline
        \noalign{\smallskip}
        \textbf{MCNet (Ours)}       & \textbf{72.33} & \textbf{67.70} & \textbf{63.50} & \textbf{60.34} & \textbf{57.59} & \textbf{54.70} & \textbf{52.13} & \textbf{50.41} & \textbf{49.08}   \\
        \noalign{\smallskip}
        \hline
        
    \end{tabular}
\end{center}
\end{table*}
\setlength{\tabcolsep}{1.4pt}

\subsection{Experimental Setup}
\textbf{Datasets.} We conduct the experiments on three benchmark datasets, i.e., CIFAR100 \cite{krizhevsky2009learning}, miniImageNet \cite{vinyals2016matching} and CUB200 \cite{wah2011caltech} to evaluate the performance of our proposed method. CIFAR100 consists of $60,000$ $32 \times 32$ images from $100$ classes, with $500$ images for training and $100$ images for testing in each class. We utilize $60$ classes for base session, while the other $40$ classes are divided into $8$ parts for incremental sessions. There are $5$ classes in each incremental session and we only choose $5$ samples from $500$ training images in each class for incremental training. MiniImageNet also contains $60,000$ images from $100$ classes, where each class has $500$ samples for training and $100$ samples for testing. The session split is same as that of CIFAR100 and there are also only $5$ samples selected for training in each novel class in incremental sessions. The images are resized to $84 \times 84$. CUB200 is a fine-grained dataset of bird species that consists of $11,788$ images of $200$ categories, where the images are resized to $224 \times 224$. We select $100$ classes for base session and the other $100$ classes for incremental sessions. Each incremental session consists of $10$ classes while there are $5$ samples in each class for training and about $30$ samples for testing. The session splits of all datasets are same as those in \cite{TOPIC,CEC,F2M}.

\textbf{Implementation Details.} Following \cite{TOPIC,CEC,F2M}, we employ ResNet18 \cite{he2016deep} as the embedding network for CUB200 and miniImageNet, and ResNet20 \cite{he2016deep} for CIFAR100. For ResNet18, we utilize the first 3 block layers as base embedding network and the last block layer as the CNN head, while the Transformer head consists of 2 attention blocks. Similar structures are applied for ResNet20, while the differences are the base embedding network contains 2 block layers and the Transformer head contains 3 attention blocks. The semantic embedding network is constructed with 2 fully connected layers with the ReLU function. We train the model with the SGD optimizer, where the learning rate is 0.05 for CIFAR100 and miniImageNet, and 0.01 for CUB200 in base session, and the batch size is set to 64 for all datasets. In the incremental sessions, the learning rate is set to 0.001 for both CIFAR100 and miniImageNet, and 0.0001 for CUB200. As for the hyper-parameters, the $\alpha$ is set to 0.4 for all three datasets and the $\lambda$ is set to 16, 8, 0.5 for CIFAR100, miniImageNet and CUB200 respectively.

\begin{table*}[t]
\begin{center}
    \caption{Comparison with the state-of-the-art methods on CUB200 dataset.}
    \label{table:cub}
    \setlength{\tabcolsep}{2mm}
	\renewcommand\arraystretch{1.1}
    \begin{tabular}{lccccccccccc}
        \hline\noalign{\smallskip}
        \multirow{2}*{\textbf{Methods}}     & \multicolumn{11}{c}{\textbf{Acc. in each session (\%)}}    \\ \cline{2-12}
        \noalign{\smallskip}
		~                          & 0 &  1  & 2 & 3 & 4 & 5 & 6 & 7 & 8 & 9 & 10      \\
        \hline
        \noalign{\smallskip}
        iCaRL \cite{icarl}      & 68.68 & 52.65 & 48.61 & 44.16 & 36.62 & 29.52 & 27.83 & 26.26 & 24.01 & 23.89 & 21.16  \\
        EEIL \cite{EEIL}       & 68.68 & 53.63 & 47.91 & 44.20 & 36.30 & 27.46 & 25.93 & 24.70 & 23.95 & 24.13 & 22.11  \\
        LUCIR \cite{LUCIR}        & 68.68 & 57.12 & 44.21 & 28.78 & 26.71 & 25.66 & 24.62 & 21.52 & 20.12 & 20.06 & 19.87  \\
        TOPIC \cite{TOPIC}      & 68.68 & 62.49 & 54.81 & 49.99 & 45.25 & 41.40 & 38.35 & 35.36 & 32.22 & 28.31 & 26.28 \\
        SKD \cite{cheraghian2021semantic}        & 68.23 & 60.45 & 55.70 & 50.45 & 45.72 & 42.90 & 40.89 & 38.77 & 36.51 & 34.87 & 32.96   \\
        SPPR \cite{SPPR}   & 68.68 & 61.85 & 57.43 & 52.68 & 50.19 & 46.88 & 44.65 & 43.07 & 40.17 & 39.63 & 37.33  \\
        SFMS \cite{cheraghian2021synthesized}        & 68.78 & 59.37 & 59.32 & 54.96 & 52.58 & 49.81 & 48.09 & 46.32 & 44.33 & 43.43 & 43.23   \\
        \noalign{\smallskip}
        \hline
        \noalign{\smallskip}
        FFSL \cite{FFSL}   & 75.63 & 71.81 & 68.16 & 64.32 & 62.61 & 60.10 & 58.82 & 58.70 & 56.45 & 56.41 & 55.82    \\
        CEC \cite{CEC}        & 75.85 & 71.94 & 68.50 & 63.50 & 62.43 & 58.27 & 57.73 & 55.81 & 54.83 & 53.52 & 52.28    \\
        F2M \cite{F2M}       & 77.13 & 73.92 & 70.27 & \textbf{66.37} & 64.34 & 61.69 & 60.52 & 59.38 & 57.15 & 56.94 & 55.89    \\
        \noalign{\smallskip}
        \hline
        \noalign{\smallskip}
        \textbf{MCNet (Ours)}       & \textbf{77.57} & \textbf{73.96} & \textbf{70.47} & {65.81} & \textbf{66.16} & \textbf{63.81} & \textbf{62.09} & \textbf{61.82} & \textbf{60.41} & \textbf{60.09} & \textbf{59.08}  \\
        \noalign{\smallskip}
        \hline
        
    \end{tabular}
\end{center}
\end{table*}

\begin{table*}[t]
\begin{center}
    \caption{Ablation study on CUB200 to analyze the effectiveness of different components in our method.}
    \label{tab:ablation}
    \setlength{\tabcolsep}{1.5mm}
	\renewcommand\arraystretch{1.2}
    \begin{tabular}{cc|cc|c|c|c|ccccccccccc}
        \hline
        \multicolumn{2}{c|}{\textbf{SWC}} & \multicolumn{2}{c|}{\textbf{TWC}}  & \multirow{2}*{\textbf{AR}} & \multirow{2}*{\textbf{FT}} & \multirow{2}*{\textbf{Branch}} & \multicolumn{10}{c}{\textbf{Acc. in each session (\%)}}    \\  \cline{8-18}
		CH & TH & CC & SC & ~ & ~ & ~ & 0 &  1  & 2 & 3 & 4 & 5 & 6 & 7 & 8 & 9 & 10  \\
        \hline
        \checkmark &  & \checkmark &  &  &  & 1 & 77.69  & 73.46 & 69.53 & 63.98 & 63.68 & 61.18 & 58.75 & 58.41 & 56.55 & 56.01 & 54.03 \\
        \checkmark &  &  & \checkmark &  &  & 1 & 77.40 & 73.36 & 68.97 & 63.15 & 62.80 & 60.55 & 58.06 & 57.21 & 55.18 & 54.84 & 53.35 \\
        \checkmark &  & \checkmark & \checkmark &  &  & 2 & 77.69 & 73.71 & 70.38 & 64.53 & 64.56 & 62.16 & 60.11 & 59.61 & 57.68 & 57.62 & 55.98  \\
        \checkmark & \checkmark & \checkmark &  &  &  & 1 & \textbf{77.92} & \textbf{73.98} & 70.28 & 65.63 & 65.02 & 62.37 & 60.43 & 59.68 & 57.59 & 56.89 & 55.54  \\
        \checkmark & \checkmark &  & \checkmark &  &  & 1 & 76.88 & 73.10 & 69.64 & 64.17 & 64.48 & 61.48 & 60.44 & 58.90 & 56.67 & 56.24 & 54.87  \\
        \checkmark & \checkmark & \checkmark & \checkmark &  &  & 2 & 77.59 & 74.02 & 70.51 & 65.68 & 65.43 & 62.87 & 61.04 & 59.90 & 58.40 & 58.07 & 56.86  \\
        \checkmark & \checkmark & \checkmark & \checkmark & \checkmark &  & 2 & 77.57 & 73.87 & 70.35 & 65.39 & 65.69 & 63.36 & 61.07 & 61.00 & 59.18 & 58.90 & 57.63  \\
        \checkmark & \checkmark & \checkmark & \checkmark & \checkmark & \checkmark & 2 & 77.57 & 73.96 & \textbf{70.47} & \textbf{65.81} & \textbf{66.16} & \textbf{63.81} & \textbf{62.09} & \textbf{61.82} & \textbf{60.41} & \textbf{60.09} & \textbf{59.08}  \\
        \hline
        
    \end{tabular}
\end{center}
\end{table*}

\subsection{Comparison with State-of-the-Art Methods}

We compare our proposed method on three datasets with three CIL methods, i.e., iCaRL \cite{icarl}, EEIL \cite{EEIL}, LUCIR \cite{LUCIR}, four trainable classifier-based FSCIL methods, i.e., TOPIC \cite{TOPIC}, SKD \cite{cheraghian2021semantic}, SPPR \cite{SPPR}, SFMS \cite{cheraghian2021synthesized}, and three prototype-based FSCIL methods, i.e., FFSL \cite{FFSL}, CEC \cite{CEC}, F2M \cite{F2M}. The experimental results are reported in Table \ref{table:cifar}, \ref{table:mini}, \ref{table:cub} respectively. Following the session split in \cite{TOPIC}, there are $8$ incremental sessions for CIFAR100 and miniImagNet, and $10$ incremental sessions for CUB200. It is observed that our method outperforms all other methods in all sessions on the three datasets, expect for the session 3 in Table \ref{table:cub}, which demonstrates that our method is more effective in alleviating the problem of catastrophic forgetting. Specifically, on the three datasets, our approach achieves the final accuracies of 50.72\%, 49.08\% and 59.08\%, outperforming the second best approach (F2M) \cite{F2M} with 1.37\%, 1.24\% and 3.19\%, respectively. 

Additionally, we have the following two observations. (1) The prototype-based methods achieve the highest performance on FSCIL, while the traditional CIL methods and the trainable classifier-based methods are unsatisfactory. We suppose the reason is that the few novel samples in each incremental session are inadequate to support the training of the classifier, while the prototype classifier could directly construct appropriate decision boundaries without any training. Our MCNet falls in the category of the prototype-based methods, but bringing further improvement and outperforms all of other approaches. (2) The MCNet performs more effectively with the increase of sessions, while the effectiveness is slight in the first three incremental sessions. It demonstrates that a single model could resist the knowledge forgetting in the early sessions, but the effect decreases in the later sessions. In contrast, it is effective to ensemble multiple models to achieve the complementation as the forgetting increases.



\subsection{Ablation Study}

We conduct ablation studies to prove the effectiveness of the main components in our proposed method. We first utilize the CNN-Head (CH) with the vanilla Cosine Classifier (CC) or the Semantic-base cosine Classifier (SC) as the baseline. Then, we gradually involve our components, including the Structure-Wise Complementation (\textbf{SWC}) that ensembles the CNN-Head (CH) and the Transformer-Head (TH), the Task-Wise Complementation (\textbf{TWC}) that combines two models trained with CC and SC respectively, the Attention Regularization (AR) between different embedding networks and Fine-Tune (FT) the model with the PSHT loss. We choose the CUB200 dataset as an example, whose results are shown in Table \ref{tab:ablation}.

Since the performance of the last session in FSCIL is the most important result, the following analysis is based on it. It can be observed that our method with all components obtains the highest performance on CUB200 dataset. Compared with CH+CC and CH+SC, the CH+CC+SC (\textbf{TWC}-CH) brings significant gains, e.g., 1.95\% against the CH+CC and 2.63\% against the CH+SC. It demonstrates that the \textbf{TWC} combining CC and SC improves the complementation between different models effectively. The performance is also improved with the \textbf{SWC}. For example, the CH+CC+TH (\textbf{SWC}+CC) brings 1.51\% performance improvement than CH+CC and the CH+SC+TH (\textbf{SWC}+SC) brings 1.52\% performance improvement than CH+SC. We can observe that the \textbf{TWC} is a little more effective than \textbf{SWC}. We suppose the reason lies in that the base embedding networks are same in \textbf{SWC} while all parameters of embedding networks are different in \textbf{TWC}. Additionally, combining the \textbf{SWC} and the \textbf{TWC} (CH+TH+CC+SC) brings 1.32\% gain compared with \textbf{SWC}+CC and 0.88\% gain compared with \textbf{TWC}+CH, which profits from the improvement of the diversity among different models. What’s more, the performance further improved by 0.77\% with the AR, which demonstrates that the diversity of different models can be improved by prompting different embedding networks focusing on different parts without compromising their discriminant ability. Finally, it is essential to fine-tune the model in the incremental sessions for addressing the FSCIL, which brings 1.45\% gains in the last session.

\begin{figure}[t]
    \centering
    \includegraphics[width=0.8\linewidth]{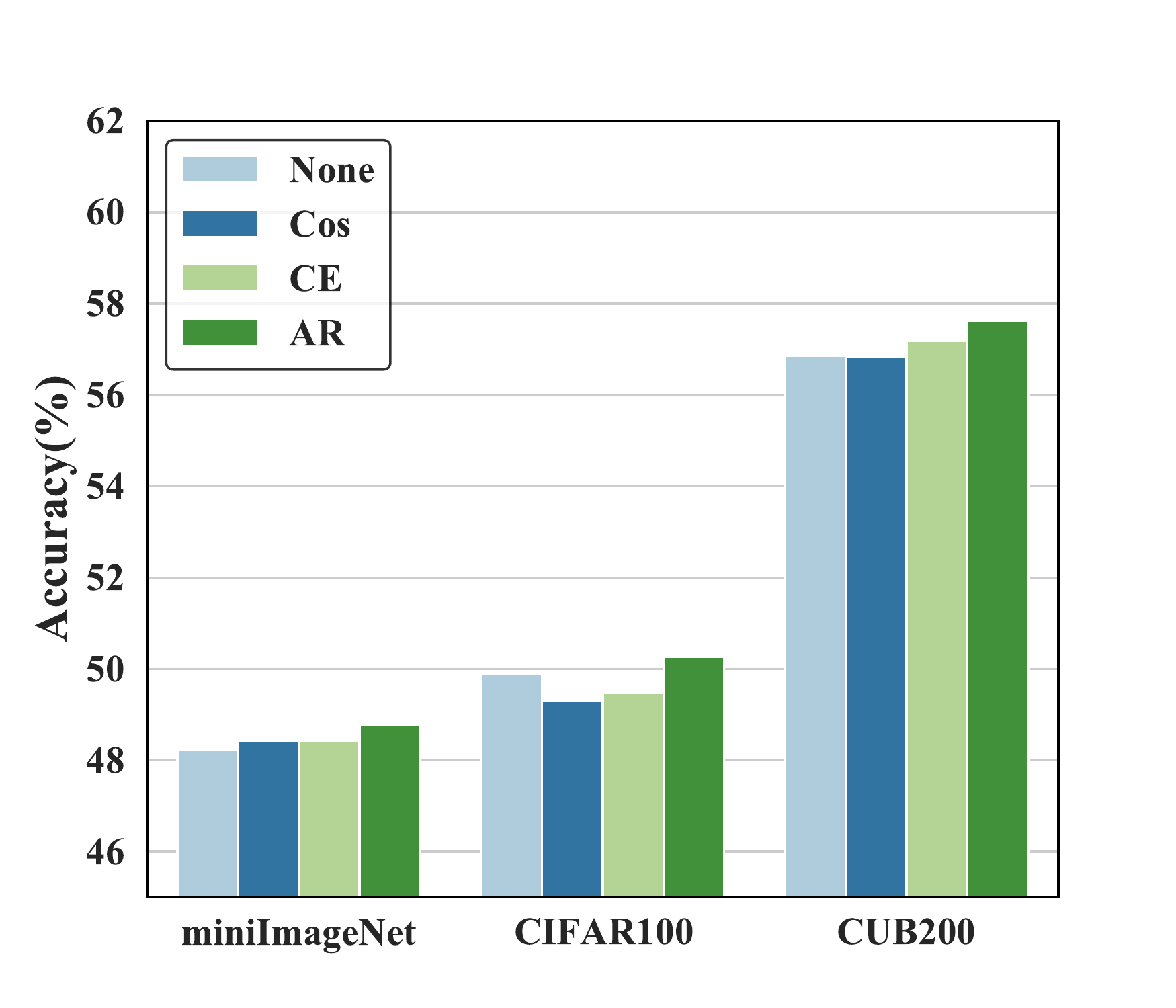}
    \caption{The impact of different regularizations on improving the model diversity.}
    \label{fig:ar}
\end{figure}

\begin{figure}[!t]
    \centering
    \includegraphics[width=0.8\linewidth]{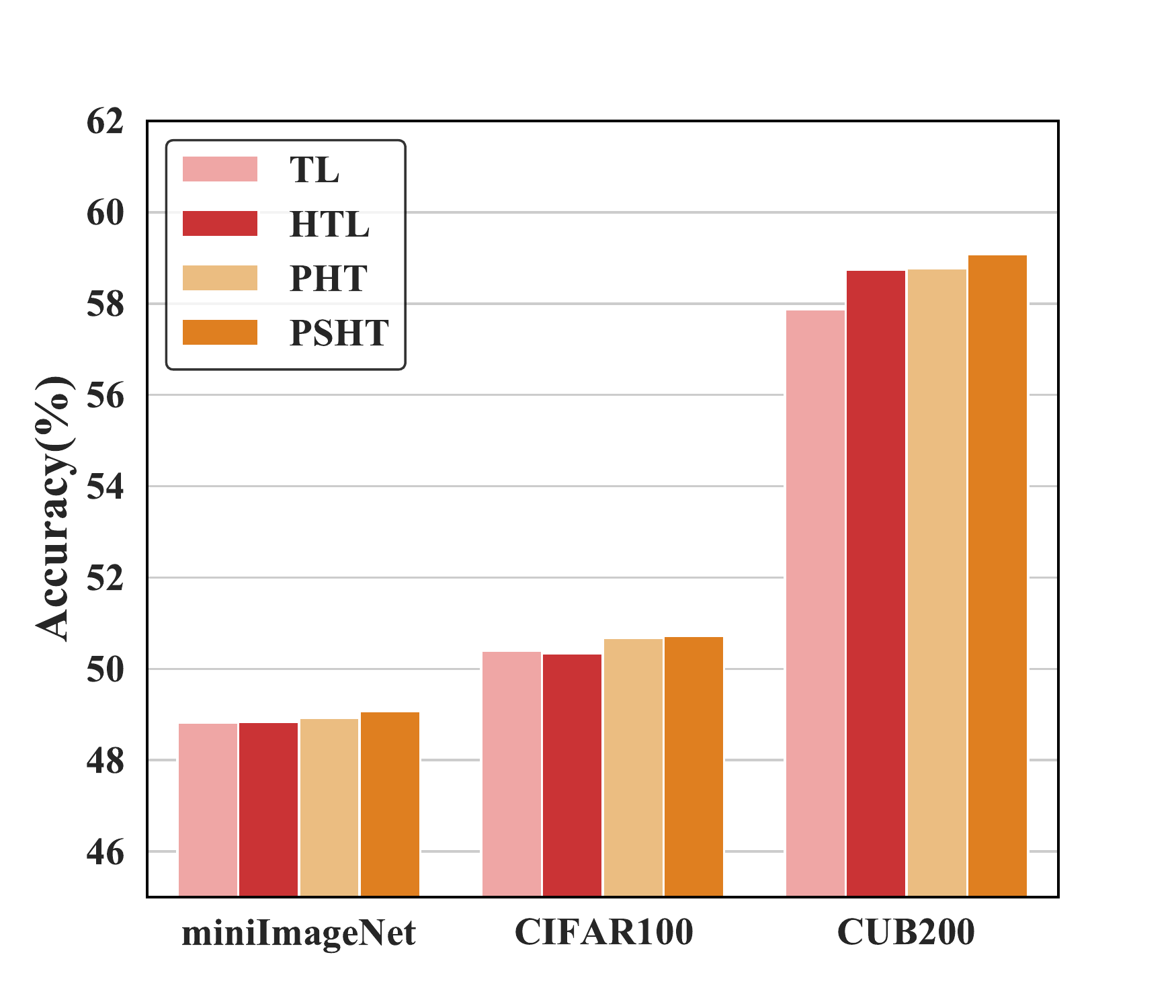}    
    \caption{The impact of different loss functions on fine-tuning the model in incremental sessions.}
    \label{fig:psht}
\end{figure}

\subsection{Further Analysis}

\textbf{Analysis of the regularization between different models.} In this section, we explore to utilize different regularization terms on the three datasets to validate the effectiveness of our proposed Attention Regularization (AR). Specifically, we employ three regularization methods: the Cosine Regularization (Cos) that reduces the cosine similarity between the feature embeddings extracted from different models, the Cross-Entropy (CE) loss that classifies the outputs of different models into different classes, and the proposed AR. The experimental results are shown in Fig. \ref{fig:ar}.

\begin{figure}[t]
    \centering
    \includegraphics[width=0.8\linewidth]{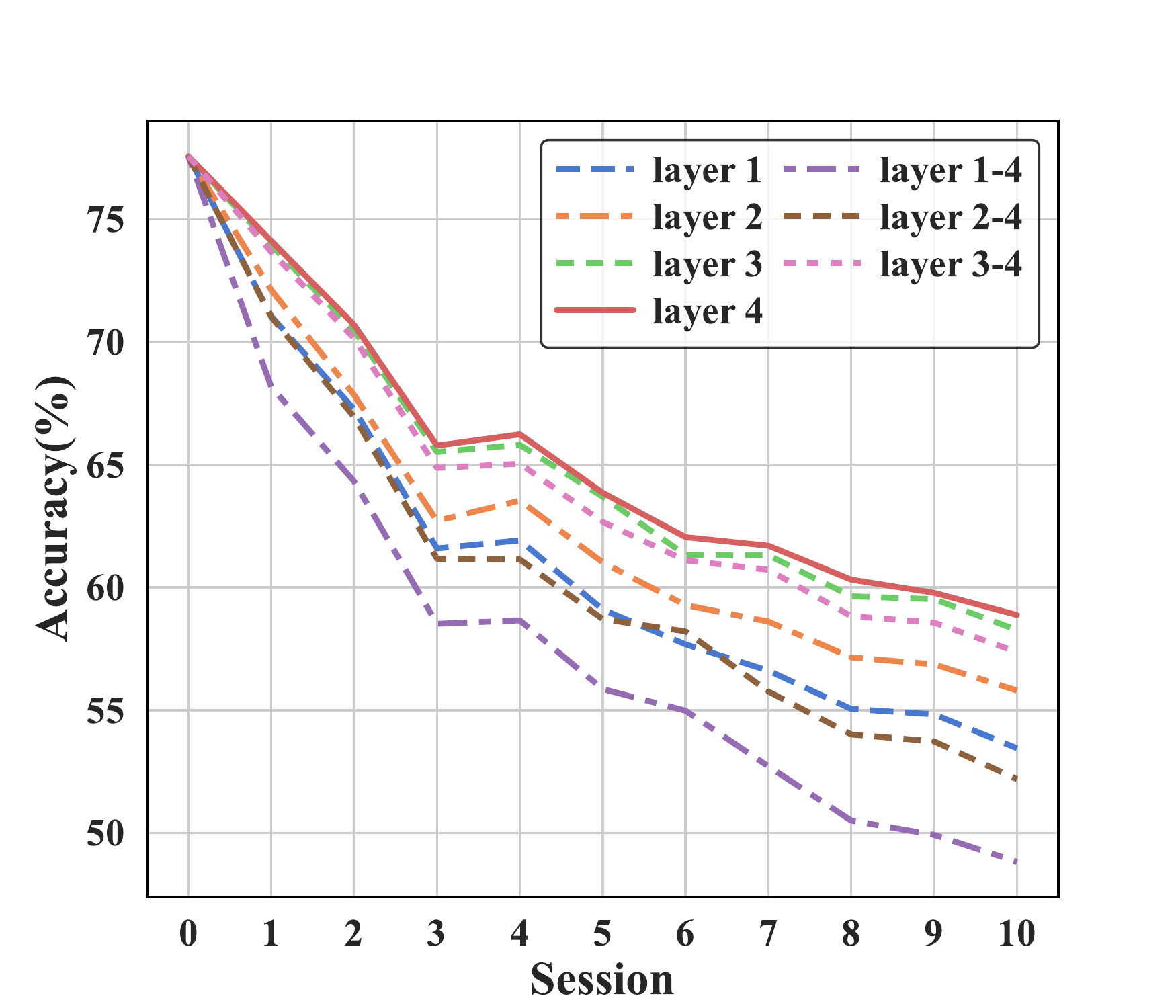}    
    \caption{Different layers fine-tuned in incremental sessions, where the layer number means which layers are updated.}
    \label{fig:layers}
\end{figure}

\begin{figure}[t]
	\centering

	\subfloat[]{\includegraphics[width=0.49\linewidth]{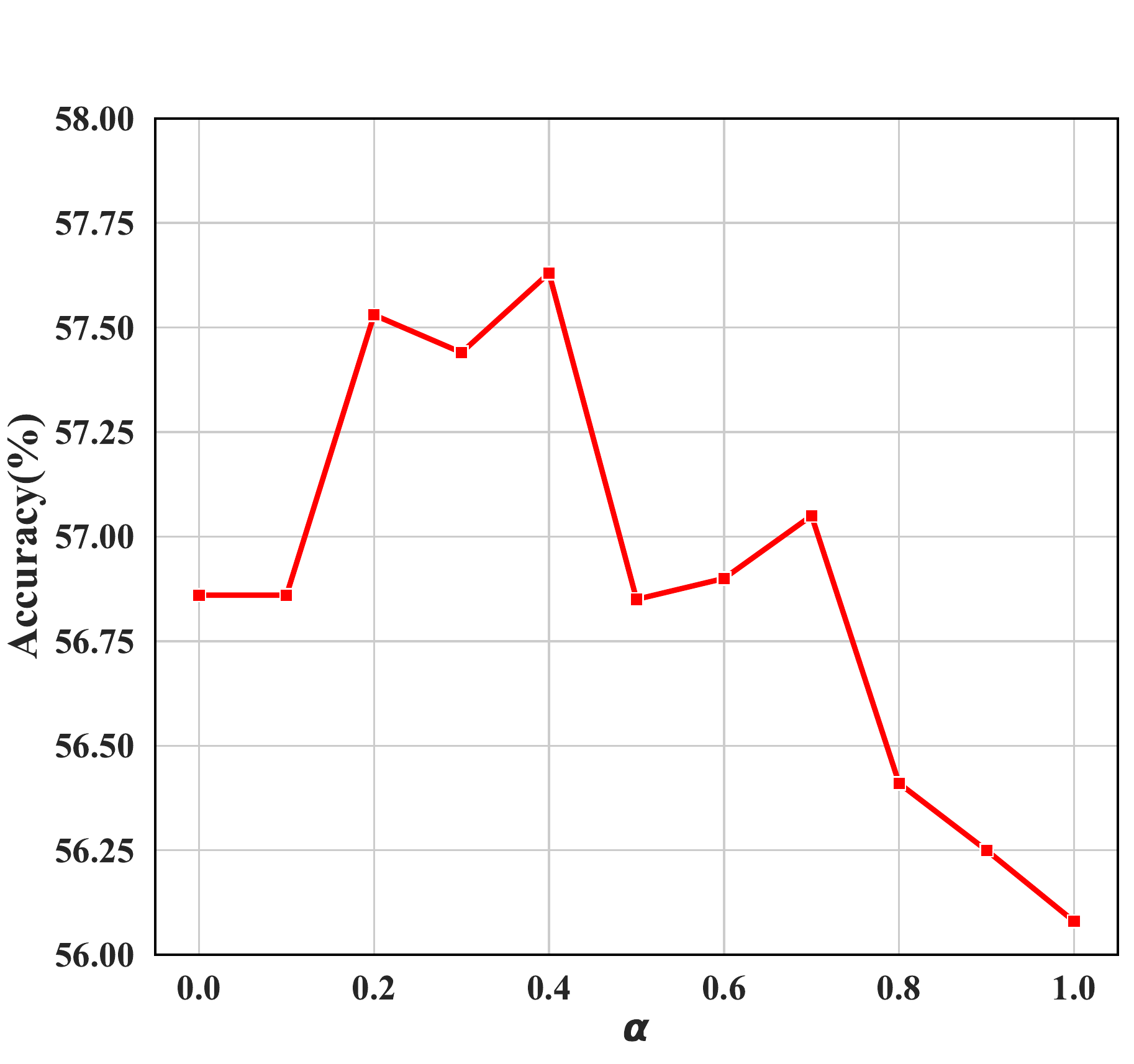}%
	\label{fig:alpha}}
	\hfil
	\subfloat[]{\includegraphics[width=0.49\linewidth]{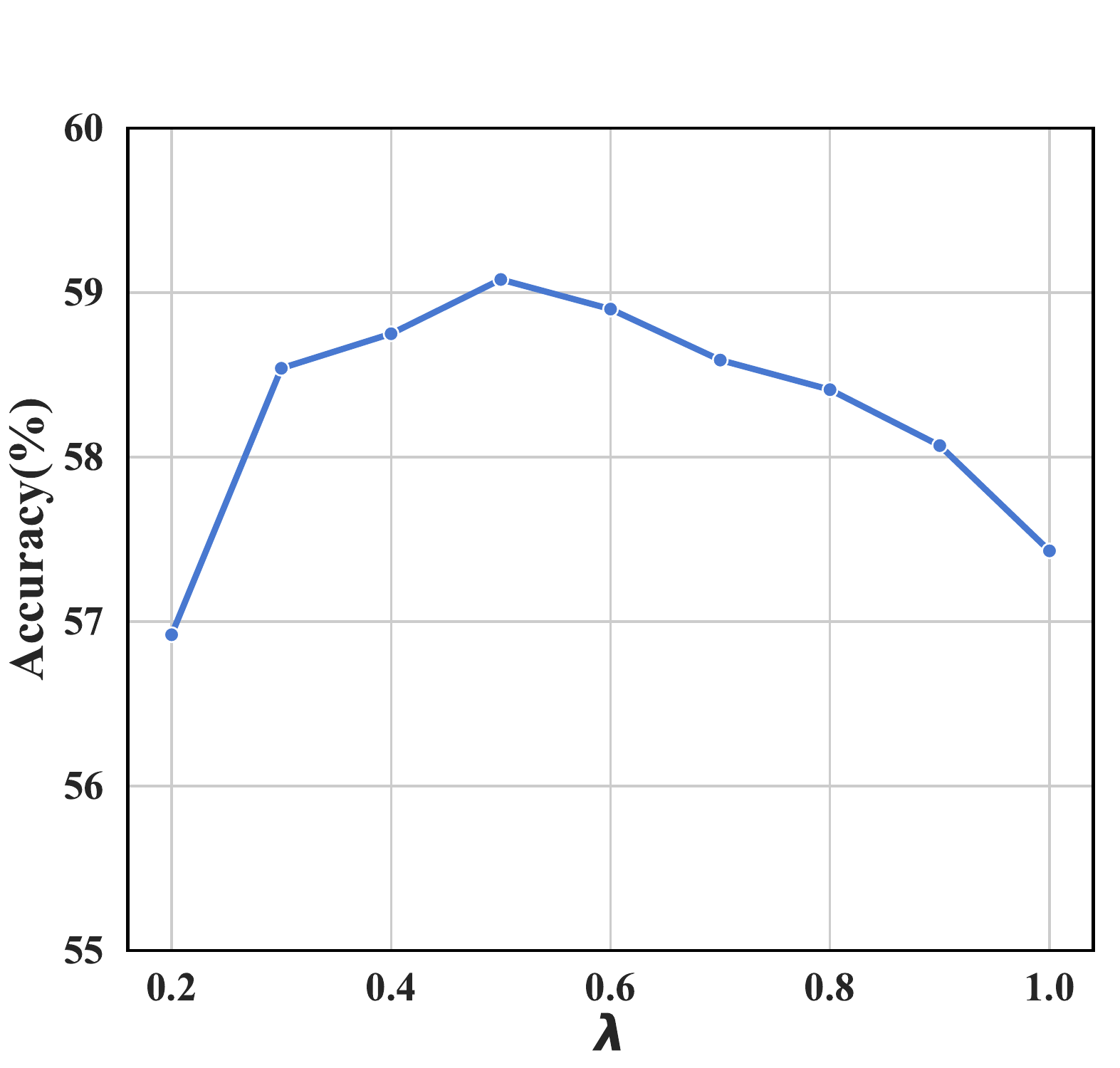}%
	\label{fig:lambda}}

	\caption{ The impact of (a) the coefficient $\alpha$, (b) the coefficient $\lambda$.}
	\label{bbb}
\end{figure}

We observe that the performances do not change significantly with the Cos or the CE on miniImageNet and CUB200, and even decrease on CIFAR100, which indicates that it may hurt the discrimination when directly increasing the difference of the output feature embeddings. By contrast, our AR brings the highest gains on our model, which demonstrates that different models focus on different parts with AR, which improves the diversity of different models effectively. 


\textbf{Analysis of the PSHT loss.} In order to prove the effectiveness of the Prototype Smoothing Hard-mining Triplet loss (PSHT), we conduct the experiments with different triplet losses, including the original Triplet Loss (TL), the Hard-mining Triplet loss (HTL), the Prototye Hard-mining Triplet loss (PHT) without smoothing and our PSHT. Figure \ref{fig:psht} reports the performance with different triplet losses. 

It can be observed that the PHT brings significant gains than the TL and HTL, which demonstrates that it is essential to push the novel samples from the old prototypes far away in embedding space to alleviate the overlap among different classes. In addition, the performance further increases with the PSHT. We suppose the reason is that the smooth prototypes represents the distribution of each category more accurately.

\begin{figure*}[!t]
	\centering

	\subfloat[]{\includegraphics[width=0.25\linewidth]{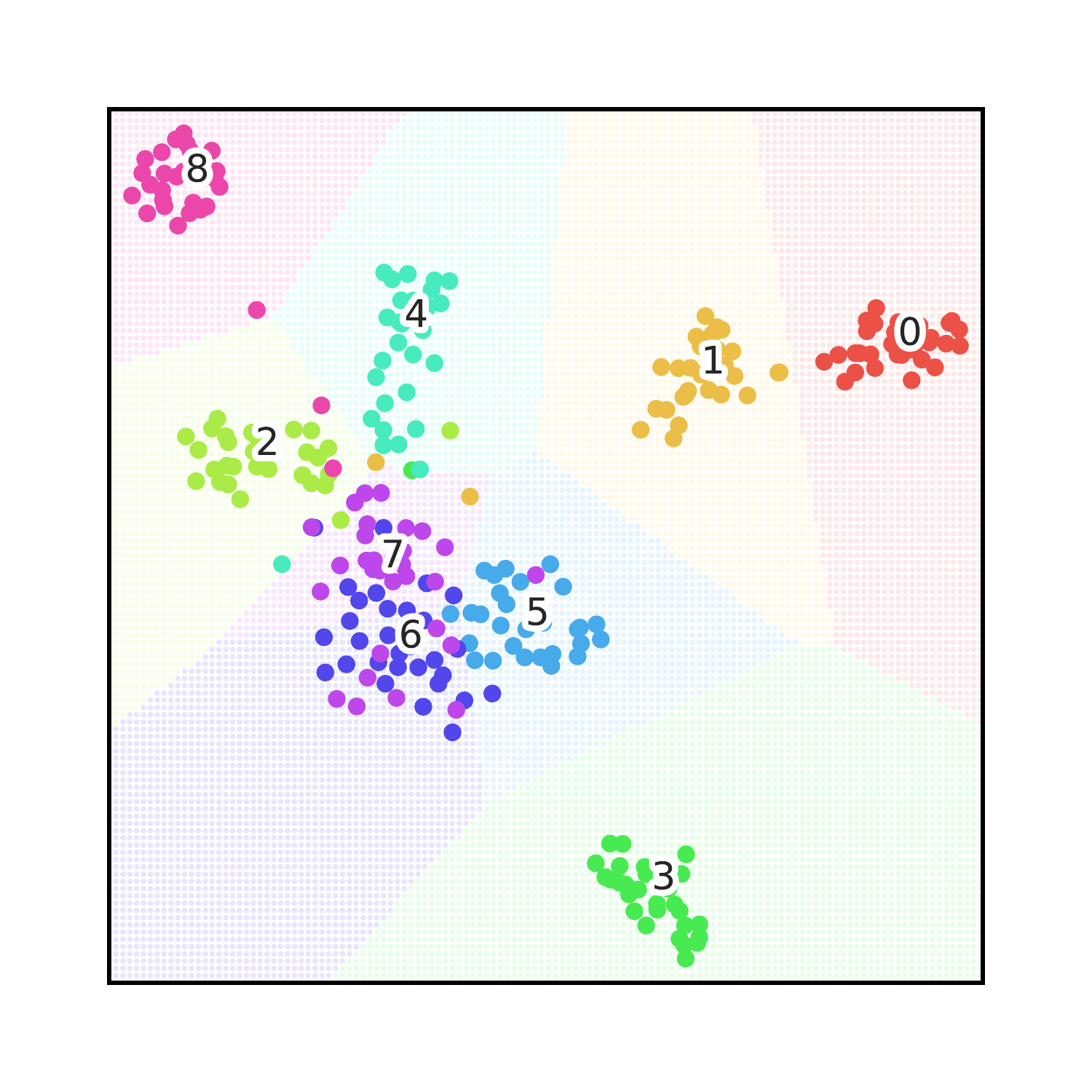}%
	\label{fig:tsne1}}
	\hfil
	\subfloat[]{\includegraphics[width=0.25\linewidth]{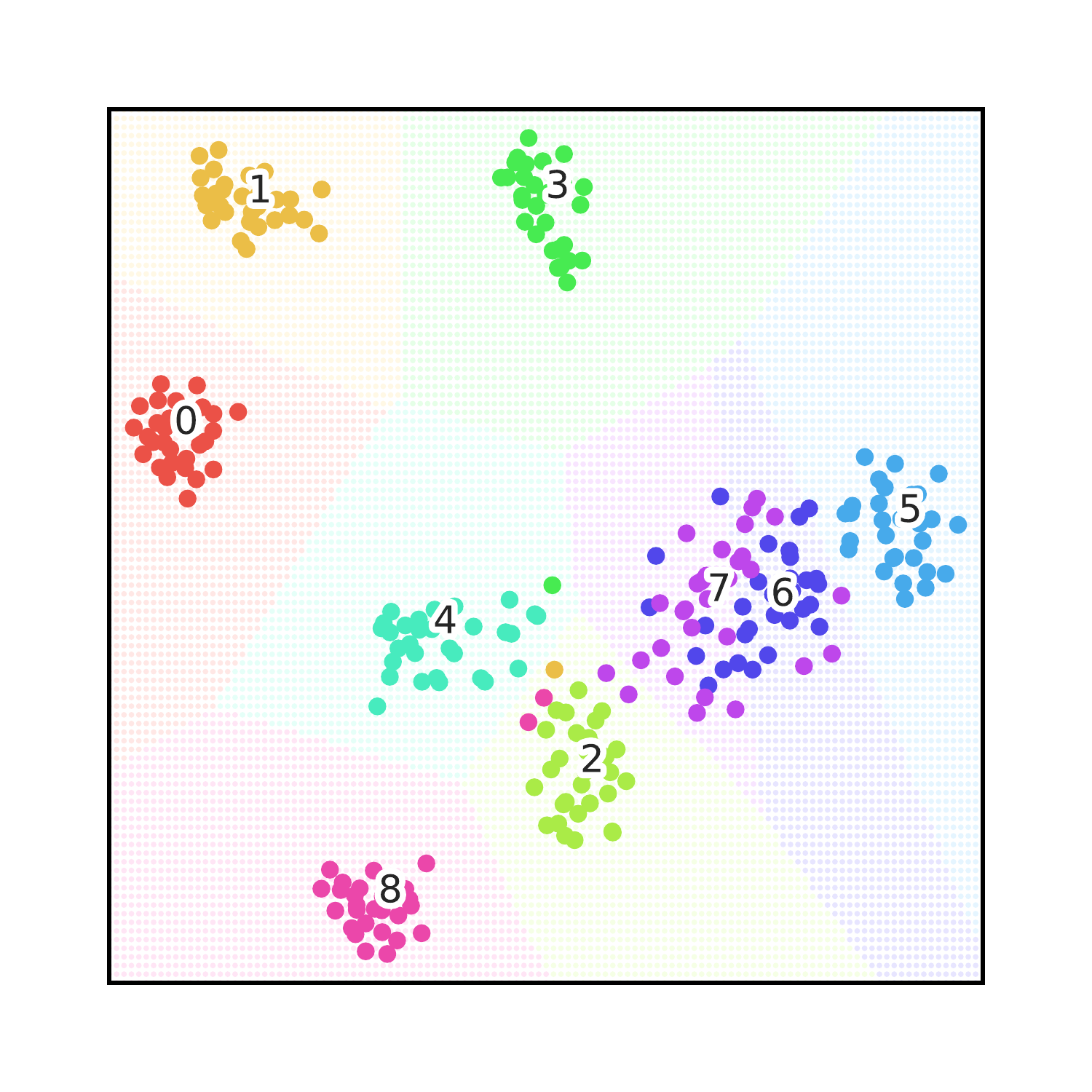}%
	\label{fig:tsne2}}
	\hfil
	\subfloat[]{\includegraphics[width=0.25\linewidth]{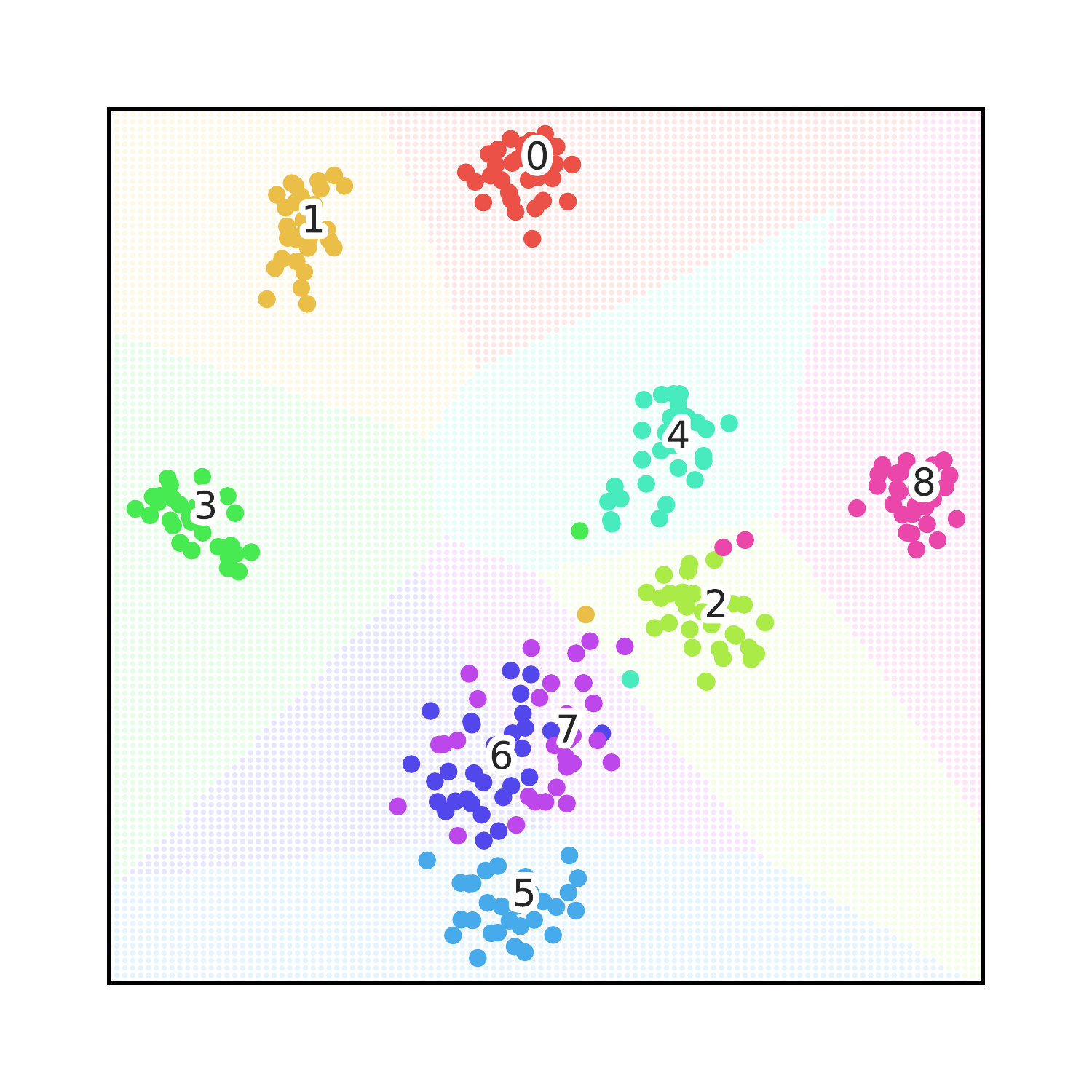}%
	\label{fig:tsne3}}

	\caption{ The t-SNE visualization of the embeddings learned by (a) a single network trained with vanilla cosine classifier, (b) the MCNet without fine-tune, and (c) the MCNet with fine-tune in incremental sessions. Class 0-4 represent the base classes while class 5-8 represent the novel classes.}
	\label{fig:tsne}
\end{figure*}

\textbf{Analysis of trainable layers in incremental sessions.} In this experiment, we update different layers on CUB200 dataset in incremental sessions and report the corresponding performance in Fig. \ref{fig:layers}. From the results we observe that only updating the last layer achieves the highest performance. Besides, we have two following observations: 1) The performance increases when we gradually fix the deeper layers. 2) When we only update one layer of the embedding network, the performance also increases with the increasing of the layer depth. These observations demonstrate that the knowledge in shallow layers is more transferable, while the deeper layers need to be updated to adapt novel concepts. 

\textbf{Analysis of the coefficient $\alpha$ in AR.} We conduct an experiment on CUB200 dataset to investigate the influence of the coefficient $\alpha$ in Eq. (\ref{Loss_base}). Figure \ref{bbb}(\subref{fig:alpha}) shows the accuracy of the last session with different $\alpha$, in which it could be observed that our MCNet achieves the best performance when $\alpha=0.4$. The impact of the AR is subtle when $\alpha$ is small since the diversity of the output features is inadequate, and the performance decreases with large $\alpha$. We suppose the reason is that it is essential to force the different embedding networks to focus on different parts, but overdoing it may hurt the discrimination of the model.

\textbf{Analysis of the coefficient $\lambda$ in incremental learning.} We further evaluate the influence of the coefficient $\lambda$ in Eq. (\ref{Loss_all}). We implement the experiment with different $\lambda$ on CUB200 dataset. The results are presented in Fig. \ref{bbb}(\subref{fig:lambda}). We observe that the performance reaches the peak when $\lambda=0.6$ and reduces as the $\lambda$ increases. It indicates that it is essential to retain the old knowledge when updating the model to alleviate the catastrophic forgetting in FSCIL. In addition, retaining too much old knowledge also affects the model to adapt to new concepts.

\textbf{Visualization.} To further prove the effectiveness of our method, we visualize the embedding space on CUB200 dataset with the t-SNE approach \cite{van2008visualizing}, as shown in Fig. \ref{fig:tsne}. We randomly select 5 classes from the base classes and 4 classes from the incremental classes respectively. We can observe from Fig. \ref{fig:tsne} (\subref{fig:tsne1}) that three novel classes are overlapped with each other and close to the base classes when there is no ensemble operation. By contrast, these feature embeddings are separated in the feature space with our proposed MCNet, shown in Fig. \ref{fig:tsne} (\subref{fig:tsne2}). Moreover, after updating the model in the incremental sessions, the novel feature embeddings are further far away from the base features, shown in  Fig. \ref{fig:tsne} (\subref{fig:tsne3}). For example, the feature embeddings of the novel class 7 are apart from the base classes 2 and 4. It indicates that our method generate better decision boundaries for both the base classes and novel classes in FSCIL.

\section{Conclusions}
This paper addresses the few-shot class-incremental learning from the perspective of complementing memorized knowledge. We have proposed a Memorizing Complementation Network (MCNet) that ensembles multiple models to realize the complementation with each other. Specifically, we employ a Structure-Wise Complementation (SWC) and a Task-Wise Complementation (TWC) respectively to improve the diversity of different models to ensure the memorized knowledge is different and complemented. We also utilize an Attention Regularization (AR) to further improve the ability of the complementation. In addition, we utilize a Prototype Smoothing Hard-mining Triplet (PSHT) loss to update only the deep layers of the model in the incremental sessions to further mitigate the catastrophic forgetting and overfitting problems. Extensive experimental results on three benchmark datasets have demonstrated the superiority of our proposed approach.

\bibliographystyle{ieeetr}
\bibliography{ref}

\begin{IEEEbiography}[{\includegraphics[width=1in,height=1.25in,clip,keepaspectratio]{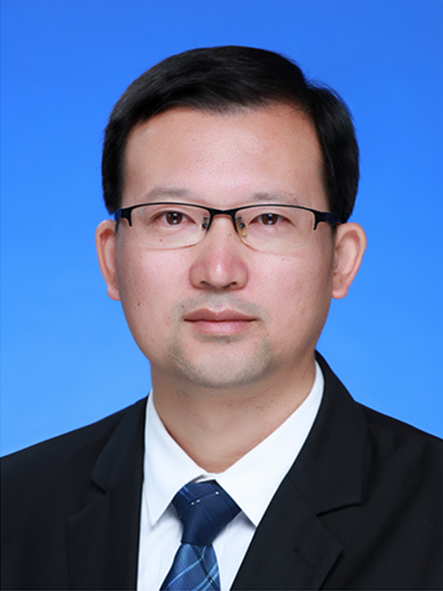}}]{Zhong Ji}
received the Ph.D. degree in signal and information processing from Tianjin University, Tianjin, China, in 2008. He is currently a Professor with the School of Electrical and Information Engineering, Tianjin University. He has authored over 100 scientific papers. His current research interests include multimedia understanding, zero/few-shot leanring, cross-modal analysis, and continual learning.
\end{IEEEbiography}

\begin{IEEEbiography}[{\includegraphics[width=1in,height=1.25in,clip,keepaspectratio]{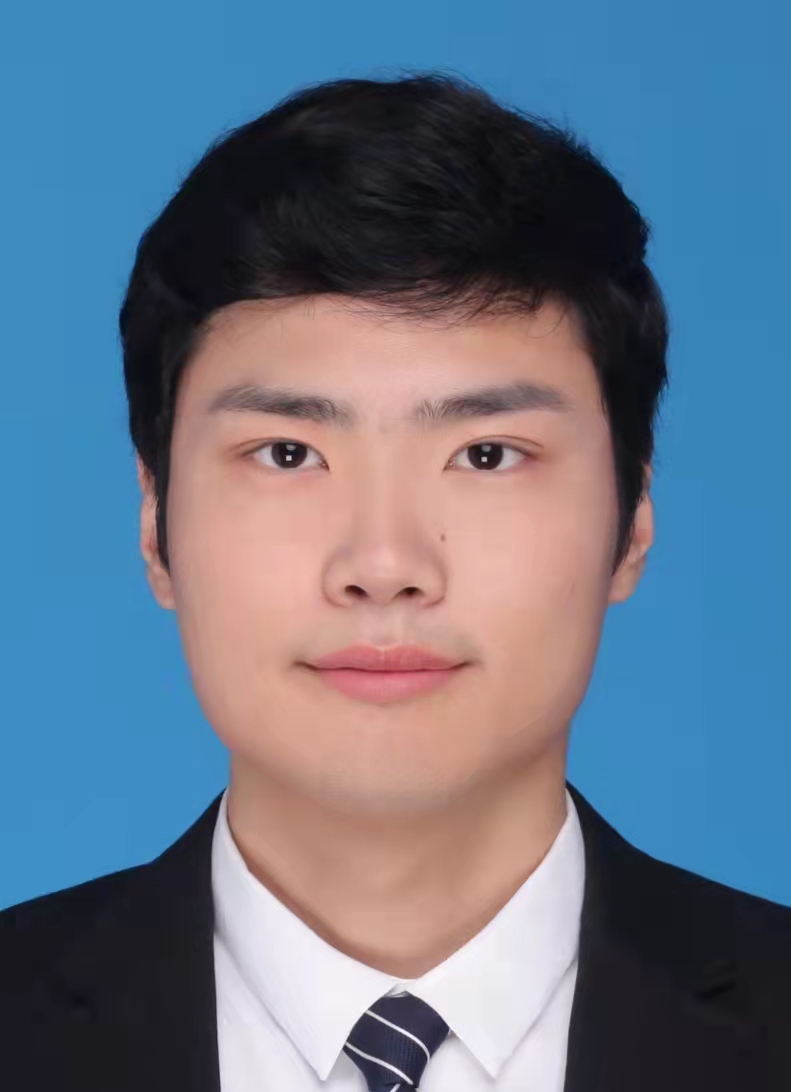}}]{Zhishen Hou}
received the B.S. degree in electronic and information engineering from Tianjin University, Tianjin, China, in 2020. He is currently pursuing a M.S. degree in the School of Electrical and Information Engineering, Tianjin University. His research interests include few-shot learning and computer vision.
\end{IEEEbiography}

\begin{IEEEbiography}[{\includegraphics[width=1in,height=1.25in,clip,keepaspectratio]{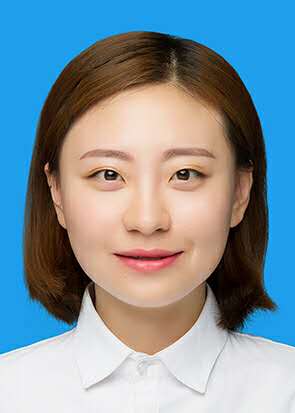}}]{Xiyao Liu}
received the B.S. degree in telecommunication engineering from Tianjin University, Tianjin, China, in 2015. She is currently pursuing a Ph.D. degree in the School of Electrical and Information Engineering, Tianjin University. Her research interests include few-shot learning, human-object interaction, and computer vision.
\end{IEEEbiography}

\begin{IEEEbiography}[{\includegraphics[width=1in,height=1.25in,clip,keepaspectratio]{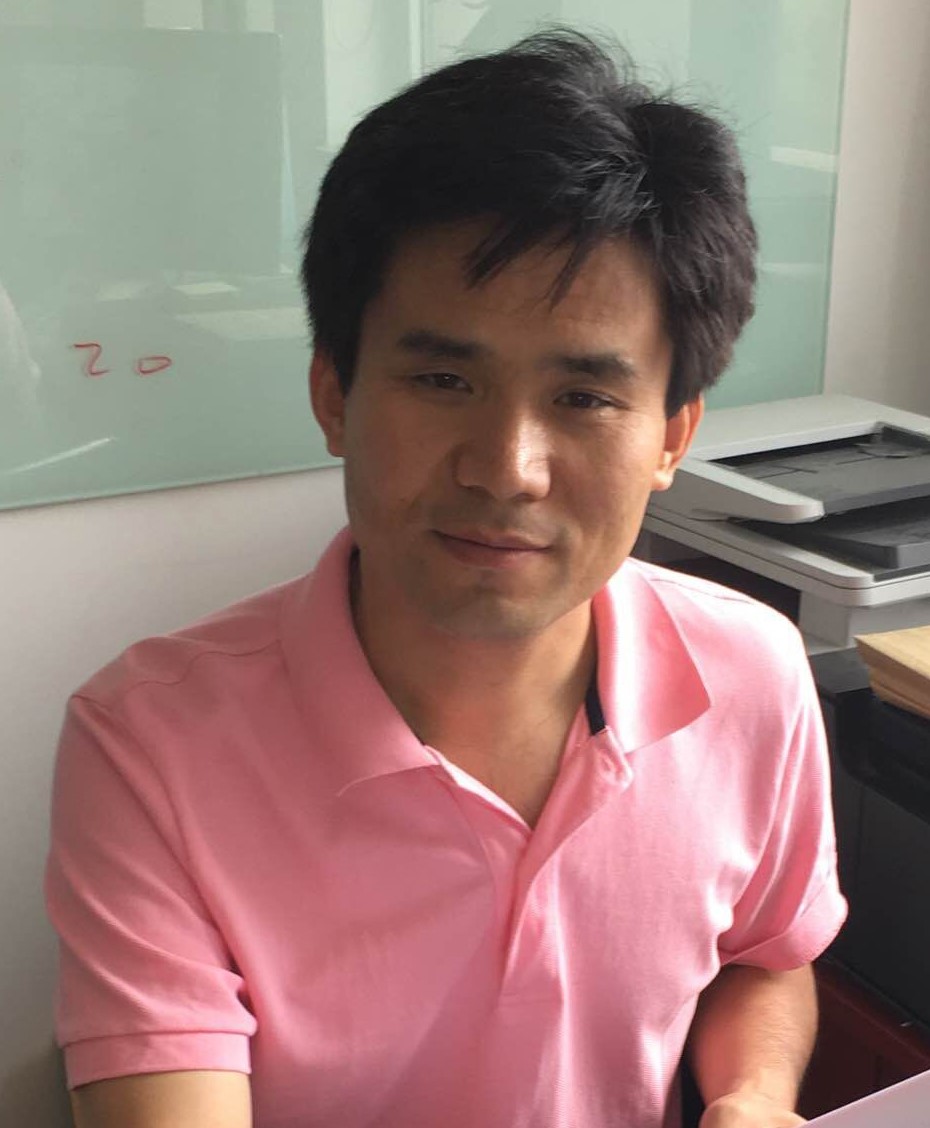}}]{Yanwei Pang}
received the Ph.D. degree in electronic engineering from the University of Science and Technology of China, Hefei, China, in 2004. He is currently a Professor with the School of Electrical and Information Engineering, Tianjin University, Tianjin, China. He has authored over 120 scientific papers. His current research interests include object detection and recognition, vision in bad weather, and computer vision.

\end{IEEEbiography}

\begin{IEEEbiographynophoto}{Xuelong Li} 
is a full professor with School of Artificial Intelligence, Optics and Electronics (iOPEN), Northwestern Polytechnical University, Xi'an 710072, P.R. China.
\end{IEEEbiographynophoto}


\end{document}